\documentclass[11pt,a4paper]{article}
\usepackage[hyperref]{acl2017}
\usepackage{times}
\usepackage{latexsym}

\usepackage{url}
\usepackage[fleqn]{amsmath}
\usepackage{multirow}
\usepackage{url}
\usepackage{amssymb}
\usepackage{multirow}
\usepackage{graphicx}
\usepackage{caption}
\usepackage{subcaption}
\usepackage{enumitem}

\aclfinalcopy 

\title{A Simple LSTM model for Transition-based Dependency Parsing}

\author{Mohab Elkaref \\
  School of Computer Science \\
  University of Birmingham, UK \\
  {\tt mxe346@cs.bham.ac.uk} \\\And
  Bernd Bohnet \\
  Google Inc. \\
  London, UK \\
  {\tt bohnetbd@google.com} \\}

\date{}

\begin{document}
\maketitle
\begin{abstract}
  We present a simple LSTM-based transition-based dependency parser. Our model is composed of a single LSTM hidden layer replacing the hidden layer in the usual feed-forward network architecture. We also propose a new initialization method that uses the pre-trained weights from a feed-forward neural network to initialize our LSTM-based model. We also show that using dropout on the input layer has a positive effect on performance. Our final parser achieves a 93.06\% unlabeled and  91.01\% labeled attachment score on the Penn Treebank. We additionally replace LSTMs with GRUs and Elman units in our model and explore the effectiveness of our initialization method on individual gates constituting all three types of RNN units.
\end{abstract}

\section{Introduction}
\label{sec:intro}

Neural Networks have become the backbone of some of the top performing transition-based dependency parsing systems. The use of a simple feed-forward network (FFN) by \newcite{chen2014fast}, kickstarted a string of improvements upon this approach. \newcite{weiss2015structured} trained a larger deeper network, and used a final structured perceptron layer on top of this. \newcite{andor2016globally} used global normalization and a large beam to achieve the state of the art results for dependency parsers of this type.

On the other hand, recurrent neural network (RNN) models have also started recieving more attention. \newcite{kiperwasser2016easy} used heirarchical tree LSTMs to model the dependency tree itself, and then passed on the extracted information to a feed-forward/output layer structure similar to that in \newcite{chen2014fast}'s original model. \newcite{dyer2015transition} used stack-LSTMs to model the current states of the different structures of a transition-based system, with separate stack-LSTMs modeling the stack, the buffer, and the history of transitions made so far.

\newcite{chen2015transition} used two Gated Recurrent Unit (GRUs) networks to represent the dependency tree, while \newcite{zhang2015top} developed TreeLSTMs to estimate the probability that a certain dependency tree is generated given a sentence.

\newcite{kiperwasser2016simple} used a conceptually simpler approach, by running a bidirectional LSTM (biLSTM) over the sentence. The inputs to these biLSTMs were various features describing the word, its part-of-speech (pos) tag, and various other structural information about each token in the sentence. The output of the biLSTMs was again passed onto a feed-forward layer to compute a hidden state before the final output layer.

These LSTM-based methods, however, all attempt to replace the original embeddings layer used by \newcite{chen2014fast} with more sophisticated feature representation but, as we pointed out, they keep the main structure of the model largely the same. That is, a hidden layer encoding the input features at the current time-step before a final output layer scores possible transitions. These hidden layers can be seen as encoding the current configuration of inputs in a manner useful only for the decision made at that point in the transition sequence. 

In contrast to these approaches, Kuncoro et. al. \shortcite{kuncoro2016dependency} extended the basic model of Chen \& Manning \shortcite{chen2014fast} by replacing the hidden layer with an LSTM, thus allowing the network to model sequences of transitions instead of only immediate input/transition pairs.

In this work we build on Kuncoro et. al.\shortcite{kuncoro2016dependency}'s approach by initialising the weights of an LSTM-based dependency parser with weights of a pre-trained Feed-Forward network. We show that this method produces a substantial improvement in accuracy scores, and is also applicable to different kinds of RNNs. An additional contribution of this paper is a refinement of the basic training model of Chen \& Manning \shortcite{chen2014fast} producing a more accurate Feed Forward model as a baseline for our experiments. 


We begin with a brief overview of transition-based dependency parsing, followed by an explanation of our baseline models; the basic FFN and LSTM-based models that are the center of this work. We then explain our proposed method for the alternative initialization of the LSTM weights, and then present the results of our experiments with a comparison with other state-of-the-art parsers. Finally we explore the use of GRUs and Elman networks in place of LSTMs, and show the effect of initializing individual gates using our proposed method on the overall performance.

\section{Transition-based Dependency Parsing}

A transition-based parsing system considers a given sentence one word at a time. The parser then makes a decision to either join this word to a word encountered previously with a dependency relation, or to store this word until it can be attached at a later point in the sentence. In this way the parser requires only a single pass through the sentence to produce a dependency tree.

In this work we use the \textbf{arc-standard} transition system \cite{Nivre:2004}, which maintains two data structures, the \emph{stack (\(S\))}, which holds the words that the parser has already seen and wishes to remember, and the \emph{buffer (\(B\))}, containing all the words that it has yet to consider, in the order in which they appear in the sentence. In addition the parser keeps a list of all dependency arcs (\(A\)) produced throughout the parse. Together, the state of the stack, the buffer, and the list of arcs are referred to as the \emph{configuration (\(x\))} of the parser. In their initial states, the buffer contains all the words of a sentence in order, with the stack containing the \emph{ROOT} token, which is typically attached to the main verb in the sentence. The parser can then perform one of 3 transitions:

\begin{itemize}[noitemsep]
\item \textbf{SHIFT} removes the front word, \(b_0\) from \(B\), and pushes it onto \(S\).
\item \textbf{LEFT-ARC} adds an arc between the top two items, \(s_0\) and \(s_1\), on \(S\) with \(s_0\) being the head. \(s_1\) is then removed from \(S\).
\item \textbf{RIGHT-ARC} adds an arc between the top two items, \(s_0\) and \(s_1\), on \(S\) with \(s_1\) being the head. \(s_0\) is then popped from \(S\).
\end{itemize}
Each of these transitions changes the state of one or more of the structures in the parser, and therefore produces a new \(x\).

\section{Baseline Models}

Our proposed approach makes use of a simple feed-forward model to improve the performance of an LSTM-based model. We show that the final network surpasses both of our baselines, which are the original feed-forward network, and an LSTM model trained with randomly initialized weights. In this section we will describe the structure of both baselines.

\subsection{Input Layer, Selected Features, \& Output Layer}
\textbf{The Embeddings layer} is a concatenation of the embedding vectors of select raw features of the parser configuration. The resulting layer is a dense feature representation of \(x\). The features used in our implementation are shown in Table \ref{tbl:feats}.
\begin{table}[h]
\fontsize{8}{12}\selectfont
\centering
\begin{tabular}{|c||l|}
\hline \bf Source & \bf Features \\ \hline
Stack & \(s_0^{w,t}, s_1^{w,t}, s_2^{w,t}\)\\
\hline
Buffer & \(b_0^{w,t}, b_1^{w,t}, b_2^{w,t}\) \\
\hline
Dependency Tree & \(rc_1(S_0)^{w,t,l}, rc_2(S_0)^{w,t,l}\)\\ 
 & \(rc_1(S_1)^{w,t,l}, rc_2(S_1)^{w,t,l}\)\\ 
 & \(lc_1(S_0)^{w,t,l}, lc_2(S_0)^{w,t,l}\)\\ 
 & \(lc_1(S_1)^{w,t,l}, lc_2(S_1)^{w,t,l}\) \\ 
 & \(rc_1(rc_1(S_0))^{w,t,l}, lc_1(lc_1(S_0))^{w,t,l}\)\\
 & \(rc_1(rc_1(S_1))^{w,t,l}, lc_1(lc_1(S_1))^{w,t,l}\)\\
\hline
\end{tabular}
\caption{\label{tbl:feats} Features extracted from a configuration. \(w\), \(t\), and \(l\) are words, pos tags, and dependency labels respectively. \(rc_n\) \& \(lc_n\) refer to the \(n^{th}\) rightmost/leftmost child.}
\end{table}

We represent the configuration of the parser at a particular timestep as a number of raw features extracted from the data structures of \(x\). We use vector embeddings to represent each of the raw features. 

Each word (\(w\)), part of speech tag (\(t\)), and arc label (\(l\)) is represented as a d-dimensional vector \(e_w \in \mathbb{R}^{d_w}\), \(e_t \in \mathbb{R}^{d_t}\), and \(e_l \in \mathbb{R}^{d_l}\) respectively. And so the embedding matrices for the different types of features are \(E^w \in \mathbb{R}^{{d_w}\times{V_w}}\), \(E^t \in \mathbb{R}^{{d_t}\times{V_t}}\), and \(E^l \in \mathbb{R}^{{d_l}\times{V_l}}\), where \(d_*\) is the dimensionality of the embedding vector for a feature type, and \(V_*\) is the vocabulary size. We add additional vectors for \emph{"ROOT"} and \emph{"NULL"} for all feature types, as well as \emph{"UNK"} (unknown), for unknown/infrequent words.

This embeddings layer is used as the input layer in all models described in this work. For all models we use dropout \cite{hinton2012improving} on the input layer. We find that this improves the final accuracy of all the networks trained.

The \textbf{output layer} \(y\) consists of nodes representing every possible transition, with one node representing Shift, and a node for every possible pair of arc transitions (Left/Right-Arc) and dependency labels. This makes the size of the output layer constant at \(2V_l+1\), regardless of the structure of the network.

\subsection{Feed-Forward Model}
\label{sec:feed-forward model}

For our FFN model we use the same basic structure of Chen and Manning \shortcite{chen2014fast} with a single hidden layer and a final softmax output layer. We however follow Weiss et. al. \shortcite{weiss2015structured} in using rectified linear units (ReLUs) \cite{nair2010rectified} as hidden neuron activation functions. Finally, we use dropout on the hidden layer similar to the input layer. The structure of the FFN is specified below.
\begin{gather*}
h = max\lbrace0, Wx + b_h\rbrace \\
y = softmax(W_hh)
\end{gather*}

Following \newcite{weiss2015structured} we set the initial bias of the hidden layer to 0.02 in order to avoid having any dead ReLUs at the start of training.

\subsection{RNN-based Model}

Our RNN-based model is an extension of the basic feed forward model, with Long Short-Term Memory (LSTM) units \cite{hochreiter1997long} standing in for the traditional feed forward hidden layers.

The change allows for the information in the parser configuration to be shared as needed with future time-steps. This lets the network at any point in the sequence of transitions make a decision based on a more informative context, that is not only based on the current configuration, or the present state of the dependency tree, but also on the changes made to them.
\begin{figure}[t]
	\captionsetup[subfigure]{justification=centering}
    \begin{subfigure}{\columnwidth}
        \includegraphics[width=\columnwidth]{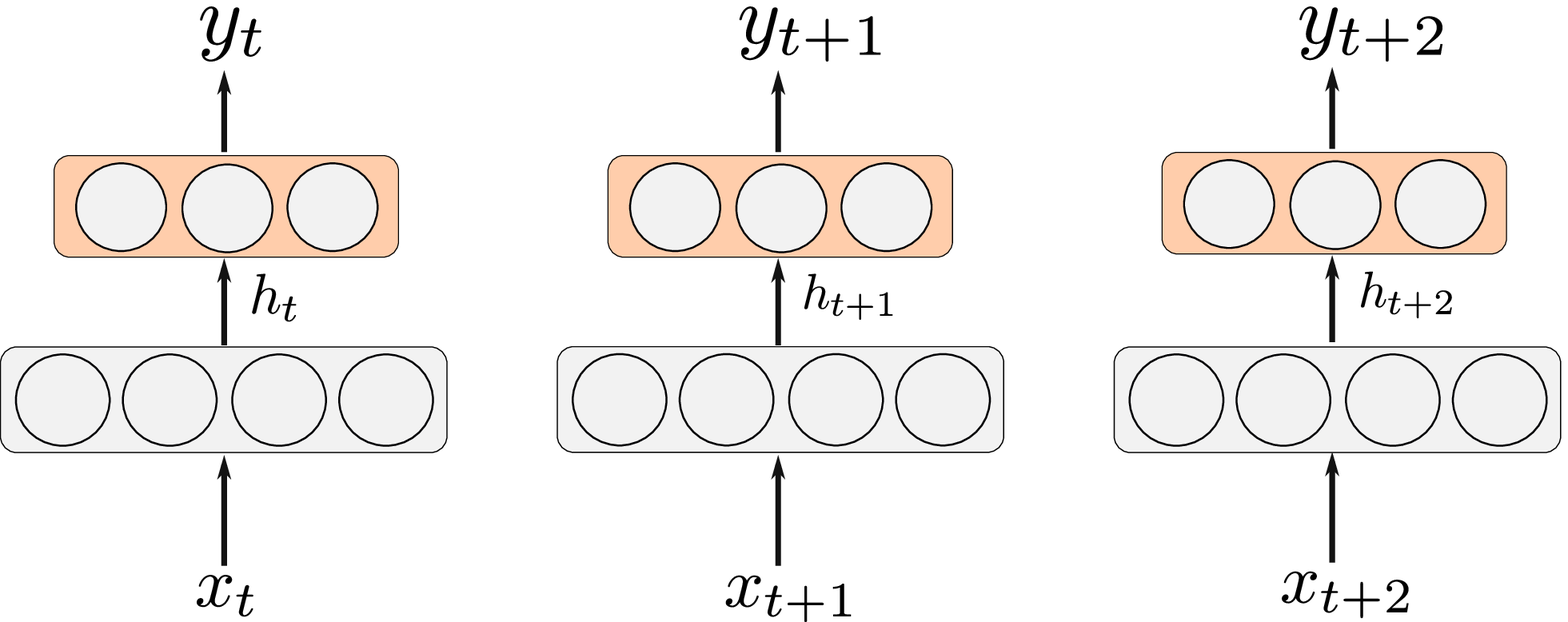}
        \caption{Feed-Forward Network (FFN)}
        \label{fig:uFFN}
        \vspace*{0.3cm}
    \end{subfigure}
    
    \begin{subfigure}{\columnwidth}
        \includegraphics[width=\columnwidth]{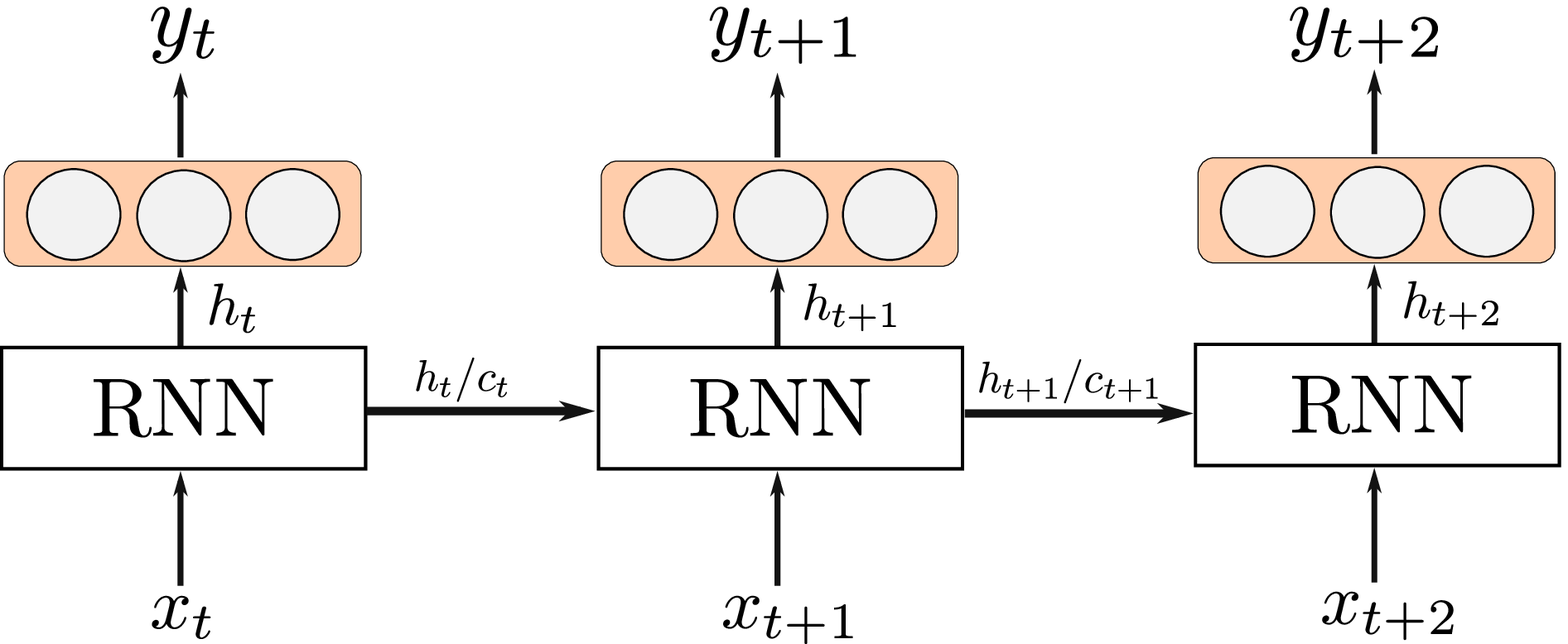}
        \caption{Unrolled RNN}
        \label{fig:uRNN}
    \end{subfigure}
    \caption{An FNN and an RNN over 3 time-steps. The FFN shown in \ref{fig:uFFN} only has access to information from the current configuration as represented in \(x\). RNNs on the otherhand also recieve information about previous configurations as encoded in the hidden states from previous time-steps. The \(h_t/c_t\) refers to the external and internal hidden states produced by an LSTM, however other types of RNN units do not necessarily maintain a \(c_t\).}\label{fig:unrolled}
\end{figure}

In their standard forms, RNNs are affected by both exploding and vanishing gradients \cite{bengio1994learning}, making them notoriously hard to train despite their expressive ability. LSTMs are a variety of RNNs that maintains an internal state \(c_t\) that forms the basis for the recurrence, and is passed from time-step to the next. This direct connection is not interrupted by any weight matrixes, as would be the case in simpler RNN architectures such as Elman networks \cite{elman1990finding}, but is instead scaled and added to by a number of gates that handle extracting and scaling information from the input data, and computing a final hidden state \(h_t\) at each time step to pass on to deeper layers. This uninterrupted connection of internal states throughout the sequence is an important part of how LSTMs address the shortcomings of RNNs.

There have been a variety of architectures in literature referred to as LSTMs, all bearing slight differences to the basic LSTM unit. The definition of the LSTM we use in this work is shown below.
\begin{gather*} 
i_t = \sigma(W_{xi}x_t + W_{hi}h_{t-1} + b_i) \\
j_t = tanh(W_{xj}x_t + W_{hj}h_{t-1} + b_j) \\
f_t = \sigma(W_{xf}x_t + W_{hf}h_{t-1} + b_f) \\
o_t = \sigma(W_{xo}x_t + W_{ho}h_{t-1} + b_o) \\
c_t = c_{t-1} \odot f_t + i_t \odot j_t \\
h_t = tanh(c_t) \odot o_t
\end{gather*}
With the final softmax output layer, just as with the FNN model.
\[y = \mathit{softmax}(W_{h}h_t)\]

Unlike Kuncoro et. al. \shortcite{kuncoro2016dependency}, we do not use peephole connections like those suggested by Graves \shortcite{graves2013generating}. Additionally, we add a bias of 1 to the LSTM's forget gate following Gers et al. \shortcite{gers2000learning}. 
Finally, we also apply a dropout similar to that in Zaremba et. al. \shortcite{zaremba2014recurrent}.

As shown in this definition, the LSTM cell maintains an internal state \(c_t\), where the previous internal state \(c_{t-1}\) is modulated at each time-step by the forget gate \(f_t\), and then added to by a scaled selection of the current input \(x_t\) by the input gates \(i_t\) and \(j_t\). This new \(c_t\) is then used for the external state \(h_t\) and passed on to the next time-step. All gates rely on weighted activations of the current input \(x_t\) and the previous external state \(h_{t-1}\).

This pair of hidden states allows the LSTM to contribute to long-term decisions with \(c_t\), while still being able to make immediate or short-term decisions with \(h_t\), and it is this final calculation of \(h_t\), along with \(o_t\), that is the focus of our contribution in this work.

\section{Initializing LSTM gates}
\label{sec:init-lstm}
\begin{figure}[!h]
	\captionsetup[subfigure]{justification=centering}
    \begin{subfigure}{\columnwidth}
        \includegraphics[width=\columnwidth]{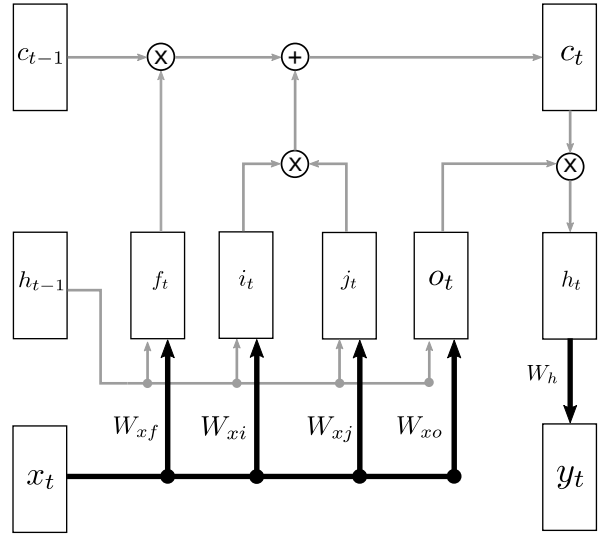}
        \caption{LSTM-based Model}
        \label{fig:LSTM}
        \vspace*{0.3cm}
    \end{subfigure}
    
    \begin{subfigure}{\columnwidth}
        \includegraphics[width=\columnwidth]{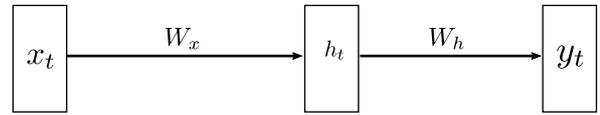}
        \caption{FNN-based Model}
        \label{fig:FFN}
    \end{subfigure}
    \caption{A comparison of the architecture of an FFN and an LSTM-based model. The bold arrows represent the weight matrices that are roughly equivalent to those in an FFN, and \(y_t\) is the final softmax layer that scores each possible transition. We only show labels for the matrices that we initialize with their FFN counterparts, \(W_x\rightarrow W_{x*}\) and \(W_{h}\rightarrow W_{h}\), where \(* \in \lbrace i, j, f, o\rbrace\). Additionally we replace the biases of the LSTM gates with the bias of the hidden layer of the FFN, \(b_h\rightarrow b_*\), and all the FFN trained embeddings for all feature types.} \label{fig:FFN-LSTM}
\end{figure}

Much has been written about the need for careful initialization of weights, often done to complement certain optimisation methods such as gradient descent with momentum in \cite{sutskever2013importance}. For deep networks, Hinton et. al. \shortcite{hinton2006fast} and later Bengio et. al. \shortcite{bengio2007greedy} approached initialization differently by using a greedy layer-wise unsupervised learning algorithm, which trains each layer sequentially, before fine-tuning the entire network as a whole.


Le et. al. \shortcite{le2015simple} suggested replacing traditional tanh units in a simple RNN with ReLUs, in addition to initializing the weights with an identity matrix.

As previously mentioned, Gers et al. \shortcite{gers2000learning} suggested initializing the bias of the forget gate \(b_f\) of an LSTM to 1. This allowed the LSTM unit to learn which information it needed to forget as opposed to detecting the opposite. This was later shown by Jozefowicz et. al. \shortcite{jozefowicz2015empirical} to improve performance of an LSTM on a variety of tasks.

Alternatively, it has become increasingly common to use tuned outputs from one network as initialization for another. For example the use of pre-trained embeddings as initialization for word vectors has become de facto standard procedure for tasks such as dependency parsing, language modelling and question answering.

Following this approach, we propose initializing the LSTM weights, specifically the \(W_{x*}\) and bias \(b_*\) of all LSTM gates, with the weight matrix \(W_x\) and hidden bias \(b_h\) of a pre-trained, similarly structured feed-forward neural network. We also initialize the embedding matrices used \(E^{w,t,l}\) and the weights of the final softmax layer \(W_{hy}\) with those of the pre-trained feed-forward network.

To illustrate this idea we reproduce a modified version of the LSTM architecture diagram appearing in \cite{jozefowicz2015empirical} in Figure \ref{fig:FFN-LSTM}, with the addition of the final softmax layer \(y_t\). The flow of information from the current input \(x_t\) to \(y_t\) (as shown by the bold arrows) is almost identical to that in an FFN, except for the addition of \(h_{t-1}\) as input to \(o\), and the ``interference" of information from \(c_t\) to produce \(h_t\).

This approach rests on the 2 hidden states of the LSTM requiring different information from the same input data. Since \(h_t\)is more concerned with immediate decisions, it would strongly benefit from the trained weights of a feed-forward network, which are tuned to extract the maximum relevant information from the input of the current time-step, since it has no access to prior information.

The various LSTM gates would still be able to learn to use information from \(h_{t-1}\) but would be in a better position to do so with the biases and input weights closer to an optimum configuration.

Moreover, the internal state \(c_t\) would receive less severe errors early on in the training process, owing to a better contribution from \(o_t\) in the calculation of \(h_t\), and a less disruptive result from \(c_t\) due to the input and forget gates initially behaving more similarly to the regular hidden layer of the original FFN.

This would mean less pressure on the weights of the input and forget gates to adapt to immediate decisions while the internal state would be more capable of gradually learning longer term patterns.

We will henceforth differentiate networks initialized in the manner described in this section by referring to them as \textbf{bootstrapped models}, while we refer to the usual randomly initialized networks as \textbf{baselines models}.

\section{Experiments}
\label{sec:exp}

We begin by comparing the performance of our FFN and LSTM baseline networks with our bootstrapped model. For all networks we ran a model with a single hidden layer 256 neurons/LSTM units wide. The embeddings dimensions used were \(d_w = d_t = d_l = 100\). \newcite{weiss2015structured} showed that large gains can be made with a grid search to tune learning hyperparameters as well embeddings sizes and hidden layer dimensions, which we did not perform due to its very high computational cost. We use the GloVe pre-trained embeddings produced by \newcite{pennington2014glove} to initialize the word vectors.

Learning is done with mini-batch stochastic gradient descent (SGD) with momentum to minimise logistic loss with the learning rate \(\alpha = 0.05\) and momentum \(\mu = 0.9\). We also use an additional \(l_2\) regularization cost (\(\lambda = 10^{-8}\)).
\[L(\theta) = -\sum_i log(y_i) + \frac{\lambda}{2}\Vert\theta\Vert^2\]
Where \(\theta\) represents all weight, biases, and embeddings matrices. We also set the dropout rate to 0.3 for the embeddings layers and hidden layer for both the baselines and bootstrapped model, and initialise all baseline weights randomly in the range [\(-0.01, 0.01\)].

For LSTM-based models we used truncated backpropagation throught time (BPTT), with a truncation limit \(\tau = 5\). This means that errors are propagated backwards to layers in previous time steps until a limit \(\tau\) is reached. In our experiments varying \(\tau\) between 5 and full back propagation had a negligible effect on the final accuracy of the networks, while using a truncation limit produced a significant speed up in training. We stress that this insignificant difference is most likely a task and architecture specific issue, and would probably be much more pronounced in other tasks and neural network set-ups.

\begin{table}
\centering
\fontsize{8}{12}\selectfont
\begin{tabular}{|l|cc|cc|}
\hline
\multicolumn{1}{|l||}{\multirow{2}{*}{ \bf Network Type}} & \multicolumn{2}{c|}{\bf Dev} & \multicolumn{2}{c|}{\bf Test} \\
\multicolumn{1}{|l||}{} & UAS & LAS & UAS & LAS \\ \hline\hline

\multicolumn{1}{|l||}{\multirow{1}{*}{\bf \emph{Feed-Forward Network}}} & & & & \\ 

\multicolumn{1}{|l||}{\multirow{1}{*}{\quad C \& M \shortcite{chen2014fast}}} & \multirow{1}{*}{92.00} & \multirow{1}{*}{89.70} & \multirow{1}{*}{91.80} & \multirow{1}{*}{89.60} \\ 
\multicolumn{1}{|l||}{\multirow{1}{*}{\quad \newcite{andor2016globally}}} & \multirow{1}{*}{92.85} & \multirow{1}{*}{90.59} & \multirow{1}{*}{92.95} & \multirow{1}{*}{91.02} \\ 
\multicolumn{1}{|l||}{\multirow{1}{*}{\quad \newcite{weiss2015structured}}} & \multirow{1}{*}{N/A} & \multirow{1}{*}{N/A} & \multirow{1}{*}{93.19} & \multirow{1}{*}{91.18} \\  \hline
\multicolumn{1}{|l||}{\multirow{1}{*}{\quad \it Our FFN baseline}} & \multirow{1}{*}{92.76} & \multirow{1}{*}{90.47} & \multirow{1}{*}{92.10} & \multirow{1}{*}{89.95} \\  \hline\hline

\multicolumn{1}{|l||}{\multirow{1}{*}{\bf \emph{LSTM Network}}} & & & &  \\ 

\multicolumn{1}{|l||}{\multirow{1}{*}{\quad \newcite{kuncoro2016dependency}}} & \multirow{1}{*}{N/A} & \multirow{1}{*}{87.8} & \multirow{1}{*}{N/A} & \multirow{1}{*}{87.5} \\ 
\multicolumn{1}{|l||}{\multirow{1}{*}{\quad \newcite{zhang2015top}}} & \multirow{1}{*}{92.66} & \multirow{1}{*}{89.14} & \multirow{1}{*}{91.99} & \multirow{1}{*}{88.69} \\ 
\multicolumn{1}{|l||}{\multirow{1}{*}{\quad K \& G \shortcite{kiperwasser2016easy}}} & \multirow{1}{*}{93.3} & \multirow{1}{*}{90.8} & \multirow{1}{*}{93.0} & \multirow{1}{*}{90.9} \\ 
\multicolumn{1}{|l||}{\multirow{1}{*}{\quad K \& G \shortcite{kiperwasser2016simple}}} & \multirow{1}{*}{N/A} & \multirow{1}{*}{N/A} & \multirow{1}{*}{93.9} & \multirow{1}{*}{91.9} \\ 
\multicolumn{1}{|l||}{\multirow{1}{*}{\quad \newcite{dyer2015transition}}} & \multirow{1}{*}{93.2} & \multirow{1}{*}{90.9} & \multirow{1}{*}{93.1} & \multirow{1}{*}{90.9} \\ \hline
\multicolumn{1}{|l||}{\multirow{1}{*}{\quad \it Our LSTM baseline}} & \multirow{1}{*}{93.23} & \multirow{1}{*}{90.94} & \multirow{1}{*}{92.77} & \multirow{1}{*}{90.64} \\ 
\multicolumn{1}{|l||}{\multirow{1}{*}{\quad \bf \emph{Our bootstrapped model}}} & \multirow{1}{*}{93.41} & \multirow{1}{*}{91.20} & \multirow{1}{*}{93.06} & \multirow{1}{*}{91.01} \\ \hline
\end{tabular}
\caption{\label{tbl:res}Final dev and test set scores on WSJ (SD). \protect\newcite{zhang2015top} do not use pre-trained word vectors for their final result. The values given for \protect\newcite{andor2016globally} and \protect\newcite{weiss2015structured} reflect only the performance of the greedy FFN models produced in their work, with other improvements made explained breifly in section \ref{sec:intro}. C \& M refers to Chen \& Manning, K \& G refers to Kiperwasser \& Goldberg.}
\end{table}

For our experiments we use the Wall Street Journal (WSJ) section from the Penn Treebank \cite{marcus1993building}. We use \S2-21 for training, \S22 for development, and \S23 for testing. We use Stanford Dependencies (SD) \cite{de2006generating} converted from constituency trees using version 3.3.0 of the converter. As is standard we use predicted POS tags for the train, dev, and test sets. We report unlabeled attachment score (UAS) and labeled attachment score (LAS), with punctuation excluded.

The results in Table \ref{tbl:res} show the effect of applying dropout on the input layer for our FFN baseline, when compared to the similarly sized \newcite{chen2014fast} model which has 200 neurons in its hidden layer. This is in addition to achieving very close dev score accuracy results with only a single 256 neuron hidden layer when compared to the significantly larger models of \newcite{weiss2015structured} with 2 layers of size 2048, and \newcite{andor2016globally} with 2 layers of size 1024 layers.

Comparing our 2 baseline models shows that the LSTM-based model performs much better than the FFN model, with an almost 0.5\% gain in dev score accuracy. Our main result is our bootstrapped model, which not only surpassed the original FFN baseline, but also the LSTM baseline. 


We note that our LSTM-baseline achieves a substantial improvement over the similar architecture of \newcite{kuncoro2016dependency}. The main differences in this case are a slightly larger model and using LSTMs \emph{without peephole connections}.

In addition, our bootstrapped model produces better results than all the mentioned feed forward models in addition to most of the LSTM-based approaches in Table \ref{tbl:res}, with the exception of \newcite{kiperwasser2016simple}, despite only having a single hidden layer of LSTM units and making no use of biLSTMs, TreeLSTMs, or Stack LSTMs.

\section{Discussion}

The results of our experiments seem to lend credence to the idea that learning short and long-term patterns separately is useful to the performance of an LSTM. To generalize this further, one could say that a sequence modelling task where a 1-to-1 relation between input/output pairs can be learned should first attempt this with an FFN, and then transfer that knowledge to an LSTM as described in section \ref{sec:init-lstm}, so sequence specific information can be further modelled.

An additional benefit of this approach is that it can be applied to previously trained FFNs and can improve any of the models that we have compared our results with in Table \ref{tbl:res}. This is also true of the LSTM-based models, where the strength of their contributions lies in their innovative approaches to feature extraction while keeping the rest of the network essentially the same.

For example, we can merge our work with that of \newcite{kiperwasser2016simple}, by first training their model; a biLSTM input layer going to a feed-forward hidden layer followed by an output layer, and then replacing the hidden layer with an LSTM initialized with the weights of that hidden layer.

Finally, our addition of applying dropout to the input layer can also be used here to further strengthen the performance of this example.

\section{Alternative Recurrent Units}

\begin{figure}[h!]
	\captionsetup[subfigure]{justification=centering}
    \begin{subfigure}{\columnwidth}
        \includegraphics[width=\columnwidth]{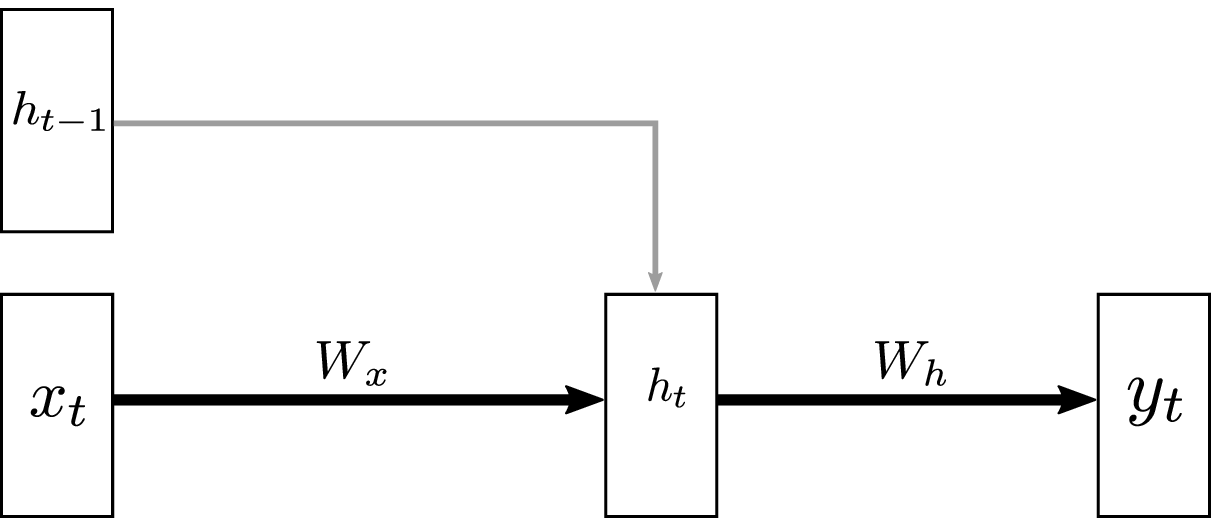}
        \caption{Elman-based Model}
        \label{fig:elman}
        \vspace*{0.3cm}
    \end{subfigure}
    
    \begin{subfigure}{\columnwidth}
        \includegraphics[width=\columnwidth]{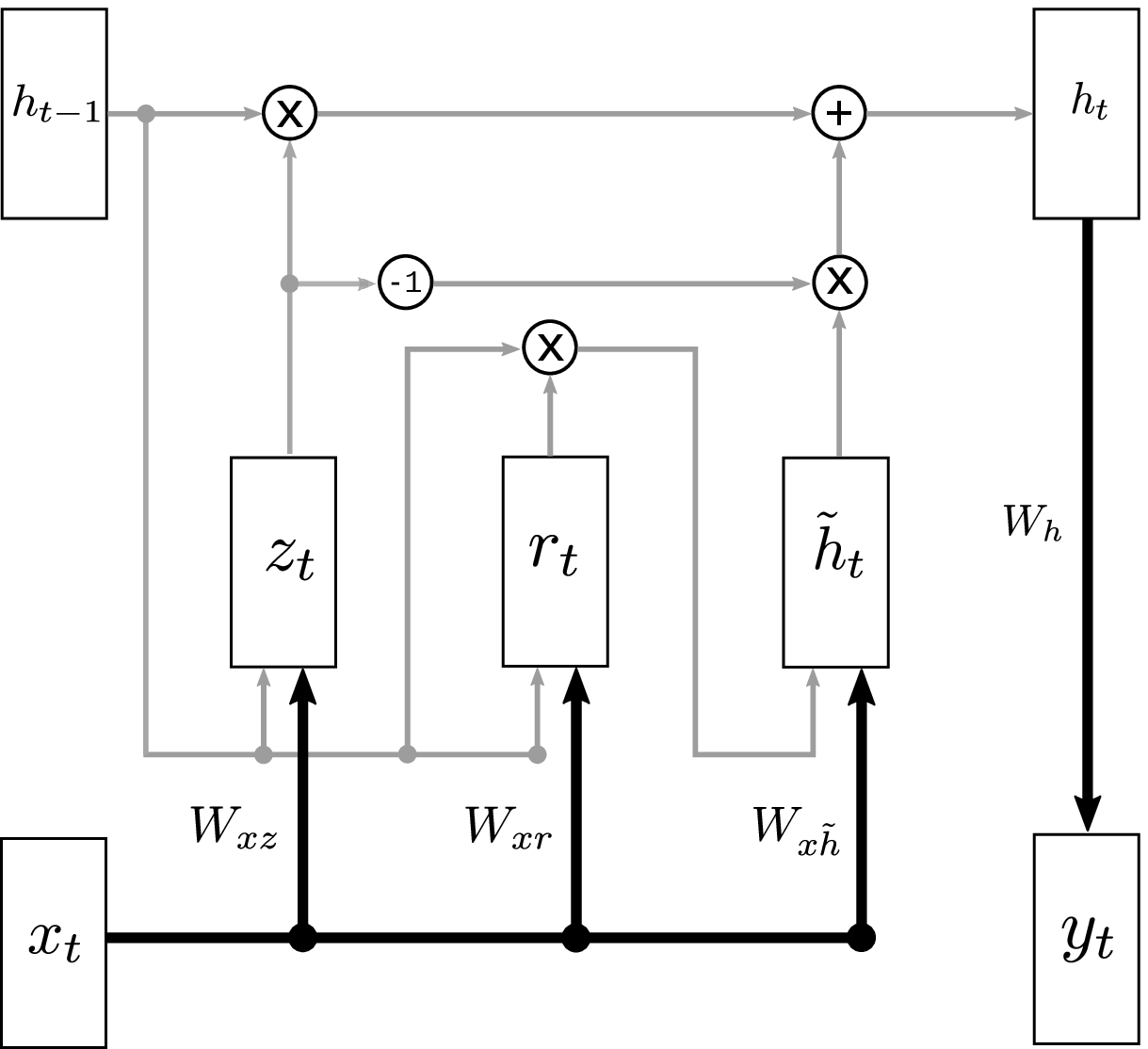}
        \caption{GRU-based Model}
        \label{fig:gru}
    \end{subfigure}
    \caption{The architectures of an Elman and a GRU-based model. As in \ref{fig:FFN-LSTM}, the bold arrows represent the path of information roughly equivalent to that in an FFN. The replaced matices in the Elman-based model are \(W_x\rightarrow W_x\), and \(W_h\rightarrow W_h\). For the GRU-based model the replaced matrices are \(W_x\rightarrow W_{x*}\), where \(* \in \lbrace z, r, \tilde{h}\rbrace\), and \(W_h\rightarrow W_h\). For both RNNs this is in addition to initializing the embeddings vectors with those trained by the baseline FFN for all feature types.}\label{fig:elman-gru}
\end{figure}

So far we have shown how to improve the performance of LSTMs by drawing parallels between the functions of certain gates and the traditional feed-forward network. In this section we attempted to do the same for 2 other popular forms of RNNs, the Simple Recurrent Network, otherwise known as the Elman network \cite{elman1990finding}, and the Gated Recurrent Unit (GRU) \cite{cho2014learning}.

\subsection{Elman networks}

The Elman network is one of the earliest and simplest RNNs found in literature. It was the subject of much study and suffered from all the original problems of vanishing and exploding gradients mentioned before, which later motivated the development and adoption of more sophisticated units such as LSTMs and GRUs.

Nevertheless there have been examples where Elman networks were capable of performing relatively well, notably the work of \newcite{mikolov2010recurrent} on language modelling and an extended memory version of Elman networks in \cite{mikolov2014learning}.

Elman networks themselves are only a simple addition to the architecture of the traditional feed-forward network. Whereas an FFN has a hidden layer, and Elman network has an additional context layer, that represents the output of the hidden layer in the previous time-step. In a way, it can be compared to output gate of an LSTM, without any additional tools to model the sequence.

In our experiment we use the ReLU activation function once more for the hidden layer similar to \newcite{le2015simple}, but without their initialization strategy. The precise definition of the Elman network that we use is shown below.
\begin{gather*}
h = max\lbrace0, W_xx + W_hh_{t-1} + b_h\rbrace \\
y = \mathit{softmax}(W_hh)
\end{gather*}

In Figure \ref{fig:elman} we illustrate the structure of this network. The simplicity of the addition here makes it far easier to draw parallels between the function of the weight matrices in the Elman network and in the FFN as shown in \ref{fig:FFN}. 

\subsection{Gated Recurrent Units}

Introduced by \newcite{cho2014learning}, GRUs are an architecture often compared to LSTMs. It also attempts to solve the gradient vanishing problem in a similar way, by keeping the modulation and addition of information in separate gates, and avoiding any weighted obstructions between the hidden states of one time-step and the next. A notable difference however is the lack of an internal state. All modifications are done directly to the external hidden state \(h_t\), potentially complicating the learning process with conflicting information about short and long-term dependencies.

Despite this apparently simpler structure, \newcite{chung2014empirical} found GRUs to outperform LSTMs on a number of tasks, and \newcite{jozefowicz2015empirical} also found that GRUs can beat LSTMs except in language modelling. However, \newcite{jozefowicz2015empirical} also found that initializing the LSTM forget gate bias \(b_f\) to 1 allowed the LSTM to almost match the performance of the GRU on other tasks.

\begin{gather*}
r_t = \sigma(W_{xr}x_t + W_{hr}h_{t-1} + b_r) \\
z_t = \sigma(W_{xz}x_t + W_{hz}h_{t-1} + b_z) \\
\tilde{h_t} = \tanh(W_{x \tilde{h}}x_t + W_{h \tilde{h}}(r_t \odot h_{t-1}) + b_h) \\
h_t = z_t \odot h_{t-1} + (1 - z_t) \odot \tilde{h_t} \\
y = \mathit{softmax}(W_{h}h_t)
\end{gather*}

The internal architecture of a GRU consists of a reset gate \(r_t\) modulating the previous state \(h_{t-1}\),  a candidate gate \(\tilde{h}\) computing the next addition to \(h_t\), and an update gate \(z_t\) controlling how much of the candidate \(\tilde{h_t}\) is added to \(h_t\).

In this case the candidate gate \(\tilde{h_t}\) is the most analogous to the hidden layer in an FFN. As shown by the bold lines in Figure \ref{fig:gru}, this flow of information appears similar to that of the output gate \(o_t\) in an LSTM, except that the additional input here is modulated by the \(r_t\) instead of receiving \(h_{t-1}\), in addition to dealing with further interference from the update gate.

\subsection{Comparison \& Results}
\label{sec:exp2}

For the experiments in this section we used the same network dimensions as in section \ref{sec:exp}, as well as the same training parameters and procedure. 

For each RNN type we trained 2 FFN and RNN baselines, one with GloVe pre-trained word embeddings\cite{pennington2014glove} and another with randomly initialized embeddings. We then trained bootstrapped models initialized with the FFN baselines. The results are shown in Table \ref{tbl:res2}.

\begin{table}[h]
\centering
\begin{tabular}{|l|cc|cc|}
\hline
\multicolumn{1}{|l||}{\multirow{1}{*}{ \bf Embeddings Type}} & UAS & LAS \\ \hline\hline

\multicolumn{1}{|l||}{\multirow{1}{*}{\bf \emph{Random Embeddings}}} & & \\ 

\multicolumn{1}{|l||}{\quad FFN baseline} & 92.21 & 89.85 \\ 
\multicolumn{1}{|l||}{\rule{0pt}{3ex}\quad LSTM baseline} & 92.16 & 89.87 \\  
\multicolumn{1}{|l||}{\quad bootstrapped LSTM} & 92.43 & 90.06 \\ 
\multicolumn{1}{|l||}{\rule{0pt}{3ex}\quad Elman baseline} & 91.97 & 89.62 \\  
\multicolumn{1}{|l||}{\quad bootstrapped Elman} & 92.40 & 90.06 \\ 
\multicolumn{1}{|l||}{\rule{0pt}{3ex}\quad GRU baseline} & 91.62 & 89.18 \\  
\multicolumn{1}{|l||}{\quad bootstrapped GRU} & 91.67 & 89.31 \\ \hline \hline 

\multicolumn{1}{|l||}{\bf \emph{Pre-trained Embeddings}} & & \\ 

\multicolumn{1}{|l||}{\quad FFN baseline} & 92.76 & 90.47 \\ 
\multicolumn{1}{|l||}{\rule{0pt}{3ex}\quad LSTM baseline} & 93.23 & 90.94 \\  
\multicolumn{1}{|l||}{\quad bootstrapped LSTM} & 93.41 & 91.20 \\ 
\multicolumn{1}{|l||}{\rule{0pt}{3ex}\quad Elman baseline} & 92.01 & 89.47 \\  
\multicolumn{1}{|l||}{\quad bootstrapped Elman} & 92.87 & 90.52 \\ 
\multicolumn{1}{|l||}{\rule{0pt}{3ex}\quad GRU baseline} & 92.21 & 89.77 \\  
\multicolumn{1}{|l||}{\quad bootstrapped GRU} & 92.14 & 89.78 \\ \hline  
\end{tabular}
\caption{\label{tbl:res2}Dev set scores on WSJ (SD) for different RNN types. The Random/Pre-trained embedding only refers to the initial word vectors of the FFN/RNN baseline. All other RNNs in these categories use the final trained embeddings of their respective FFN baseline.}
\end{table}

As in section \ref{sec:exp}, this initialization method shows a positive effect on an LSTM-based model, again surpassing both its baselines. The Elman network is stronger than expected and benefits greatly from this approach. Indeed, the bootstrapped Elman model is comparable in accuracy to some of the results in Table \ref{tbl:res}.

This cannot be said of GRUs, however, where its baselines perform significantly worse than other RNNs. Moreover, bootstrapped GRU models perform even worse than their baselines, even failing to match the accuracy of the FFNs used to initialize them. This disparity in accuracy compared to LSTMs seems to lend credence to our earlier hypothesis that learning long-term sequences can interfere with learning to make immediate decisions based on the input from the current time step. The architecture of an LSTM which maintains a long-term internal state \(c_t\) separate from a short-term external state \(h_t\), and the additional improvement gained from learning these separately, as opposed to the single common hidden state \(h_t\) in GRUs appears to provide a distinct advantage here.

The improvement achieved by a bootstrapped Elman model can thus be explained by the fact that it suffers from gradient vanishing \cite{bengio1994learning}, and so sequence specific information does not affect training to the extent that it does in GRUs.

\section{Initializing Individual Gates}

Our final set of experiments is to investigate whether or not individual gates of LSTMs and GRUs can benefit from this initialization technique. We follow the same initialization and training procedures described previously, and for every gate we also initialize its corresponding bias vectors. We keep the same size and parameters as in section \ref{sec:exp2}, and also train baselines with and without pre-trained embeddings.


\begin{table}[t]
\centering
\begin{tabular}{|l|cc|cc|}
\hline
\multicolumn{1}{|l||}{\multirow{1}{*}{ \bf Embeddings Type}} & UAS & LAS \\ \hline\hline

\multicolumn{1}{|l||}{\multirow{1}{*}{\bf \emph{Random Embeddings}}} & & \\ 

\multicolumn{1}{|l||}{\quad FFN baseline} & 92.21 & 89.85 \\ 
\multicolumn{1}{|l||}{\rule{0pt}{3ex}\quad LSTM baseline} & 92.16 & 89.87 \\  
\multicolumn{1}{|l||}{\rule{0pt}{3ex}\quad bootstrapped \(i\) gate} & 92.29 & 90.00 \\ 
\multicolumn{1}{|l||}{\quad bootstrapped \(j\) gate} & 92.38 & 89.96 \\  
\multicolumn{1}{|l||}{\quad bootstrapped \(f\) gate} & 92.25 & 89.81 \\ 
\multicolumn{1}{|l||}{\quad bootstrapped \(o\) gate} & 92.43 & 90.06\rule[-2ex]{0pt}{0pt} \\ 
\multicolumn{1}{|l||}{\quad bootstrapped all gates} & 92.38 & 90.01\rule[-2ex]{0pt}{0pt} \\ \hline \hline 

\multicolumn{1}{|l||}{\bf \emph{Pre-trained Embeddings}} & & \\ 

\multicolumn{1}{|l||}{\quad FFN baseline} & 92.76 & 90.47 \\ 
\multicolumn{1}{|l||}{\rule{0pt}{3ex}\quad LSTM baseline} & 93.23 & 90.94 \\  
\multicolumn{1}{|l||}{\rule{0pt}{3ex}\quad bootstrapped \(i\) gate} & 93.20 & 90.96 \\ 
\multicolumn{1}{|l||}{\quad bootstrapped \(j\) gate} & 93.30 & 91.02 \\  
\multicolumn{1}{|l||}{\quad bootstrapped \(f\) gate} & 93.42 & 91.22 \\ 
\multicolumn{1}{|l||}{\quad bootstrapped \(o\) gate} & 93.35 & 91.11\rule[-2ex]{0pt}{0pt} \\ 
\multicolumn{1}{|l||}{\quad bootstrapped all gates} & 93.41 & 91.20\rule[-2ex]{0pt}{0pt} \\ \hline 
\end{tabular}
\caption{\label{tbl:resLSTM}Dev set scores on WSJ (SD) for individually bootstrapped LSTM gates}
\end{table}

Bootstrapping individual LSTM gates produces mixed results, especially when considering the difference in performance between the random and pre-trained embeddings experiments.

Full bootstrapping, bootstrapping the \(j\) gate or bootstrapping the \(o\) gate seem to be the most reliable options based on these results.


\begin{table}[t]
\centering
\begin{tabular}{|l|cc|cc|}
\hline
\multicolumn{1}{|l||}{\multirow{1}{*}{ \bf Embeddings Type}} & UAS & LAS \\ \hline\hline

\multicolumn{1}{|l||}{\multirow{1}{*}{\bf \emph{Random Embeddings}}} & & \\ 

\multicolumn{1}{|l||}{\quad FFN baseline} & 92.21 & 89.85 \\ 
\multicolumn{1}{|l||}{\rule{0pt}{3ex}\quad GRU baseline} & 91.62 & 89.18 \\  
\multicolumn{1}{|l||}{\rule{0pt}{3ex}\quad bootstrapped \(r\) gate} & 91.70 & 89.15 \\ 
\multicolumn{1}{|l||}{\quad bootstrapped \(z\) gate} & 90.59 & 87.90 \\  
\multicolumn{1}{|l||}{\quad bootstrapped \(\tilde{h}\) gate} & 91.67 & 89.31\rule[-2ex]{0pt}{0pt} \\
\multicolumn{1}{|l||}{\quad bootstrapped all gates} & 91.73 & 89.20\rule[-2ex]{0pt}{0pt} \\ \hline \hline 

\multicolumn{1}{|l||}{\bf \emph{Pre-trained Embeddings}} & & \\ 

\multicolumn{1}{|l||}{\quad FFN baseline} & 92.76 & 90.47 \\ 
\multicolumn{1}{|l||}{\rule{0pt}{3ex}\quad GRU baseline} & 92.21 & 89.77 \\  
\multicolumn{1}{|l||}{\rule{0pt}{3ex}\quad bootstrapped \(r\) gate} & 92.22 & 89.79 \\ 
\multicolumn{1}{|l||}{\quad bootstrapped \(z\) gate} & 92.62 & 90.37 \\  
\multicolumn{1}{|l||}{\quad bootstrapped \(\tilde{h}\) gate} & 92.14 & 89.78\rule[-2ex]{0pt}{0pt} \\
\multicolumn{1}{|l||}{\quad bootstrapped all gates} & 89.30 & 86.09\rule[-2ex]{0pt}{0pt} \\ \hline 
\end{tabular}
\caption{\label{tbl:resLSTM}Dev set scores on WSJ (SD) for individually bootstrapped GRU gates}
\end{table}

Results for bootstrapping individual GRU gates vary drastically, with individual gates performing very differently in their random and pre-trained embedding experiments. 

Surprisingly, bootstrapping all GRU gates achieves better results than the GRU baseline for random embeddings, while severely hurting accuracy with pre-trained embeddings. All GRU experiments, bootstrapped or not, still do not perform better than the FFN baseline.


\section{Conclusion}

In this paper we have presented a simple and effective LSTM transition-based dependency parser. Its performance rivals that of far more complicated approaches, while still being capable of integrating with minimal changes to their architecture.

Additionally, we showed that the application of dropout to the input layer can improve the performance of a network. Like our other contributions here this is simple to apply to other models and is not only limited to the architectures presented in this work.

Finally, we proposed a method of using pre-trained FFNs as initializations for an RNN-based model. We showed that this approach can produce gains in accuracy for both LSTMs and Elman networks, with the final LSTM model surpassing or matching most state-of-the-art LSTM-based models.

This initialization method can potentially be applied to any LSTM-based task, where a 1-to-1 relation between inputs can first be modelled using an FFN. Exploring the effects of this method on other tasks is left for future work.

\bibliography{initRnnFF}

\begin{thebibliography}{}
\expandafter\ifx\csname natexlab\endcsname\relax\def\natexlab#1{#1}\fi

\bibitem[{Andor et~al.(2016)Andor, Alberti, Weiss, Severyn, Presta, Ganchev,
  Petrov, and Collins}]{andor2016globally}
Daniel Andor, Chris Alberti, David Weiss, Aliaksei Severyn, Alessandro Presta,
  Kuzman Ganchev, Slav Petrov, and Michael Collins. 2016.
\newblock Globally normalized transition-based neural networks.
\newblock {\em arXiv preprint arXiv:1603.06042\/} .

\bibitem[{Bengio et~al.(2007)Bengio, Lamblin, Popovici, Larochelle
  et~al.}]{bengio2007greedy}
Yoshua Bengio, Pascal Lamblin, Dan Popovici, Hugo Larochelle, et~al. 2007.
\newblock Greedy layer-wise training of deep networks.
\newblock {\em Advances in neural information processing systems\/} 19:153.

\bibitem[{Bengio et~al.(1994)Bengio, Simard, and Frasconi}]{bengio1994learning}
Yoshua Bengio, Patrice Simard, and Paolo Frasconi. 1994.
\newblock Learning long-term dependencies with gradient descent is difficult.
\newblock {\em IEEE transactions on neural networks\/} 5(2):157--166.

\bibitem[{Chen and Manning(2014)}]{chen2014fast}
Danqi Chen and Christopher~D Manning. 2014.
\newblock A fast and accurate dependency parser using neural networks.
\newblock In {\em EMNLP\/}. pages 740--750.

\bibitem[{Chen et~al.(2015)Chen, Zhou, Zhu, Qiu, and
  Huang}]{chen2015transition}
Xinchi Chen, Yaqian Zhou, Chenxi Zhu, Xipeng Qiu, and Xuanjing Huang. 2015.
\newblock Transition-based dependency parsing using two heterogeneous gated
  recursive neural networks.
\newblock In {\em EMNLP\/}. pages 1879--1889.

\bibitem[{Cho et~al.(2014)Cho, Van~Merri{\"e}nboer, Gulcehre, Bahdanau,
  Bougares, Schwenk, and Bengio}]{cho2014learning}
Kyunghyun Cho, Bart Van~Merri{\"e}nboer, Caglar Gulcehre, Dzmitry Bahdanau,
  Fethi Bougares, Holger Schwenk, and Yoshua Bengio. 2014.
\newblock Learning phrase representations using rnn encoder-decoder for
  statistical machine translation.
\newblock {\em arXiv preprint arXiv:1406.1078\/} .

\bibitem[{Chung et~al.(2014)Chung, Gulcehre, Cho, and
  Bengio}]{chung2014empirical}
Junyoung Chung, Caglar Gulcehre, KyungHyun Cho, and Yoshua Bengio. 2014.
\newblock Empirical evaluation of gated recurrent neural networks on sequence
  modeling.
\newblock {\em arXiv preprint arXiv:1412.3555\/} .

\bibitem[{De~Marneffe et~al.(2006)De~Marneffe, MacCartney, Manning
  et~al.}]{de2006generating}
Marie-Catherine De~Marneffe, Bill MacCartney, Christopher~D Manning, et~al.
  2006.
\newblock Generating typed dependency parses from phrase structure parses.
\newblock In {\em Proceedings of LREC\/}. volume~6, pages 449--454.

\bibitem[{Dyer et~al.(2015)Dyer, Ballesteros, Ling, Matthews, and
  Smith}]{dyer2015transition}
Chris Dyer, Miguel Ballesteros, Wang Ling, Austin Matthews, and Noah~A Smith.
  2015.
\newblock Transition-based dependency parsing with stack long short-term
  memory.
\newblock {\em arXiv preprint arXiv:1505.08075\/} .

\bibitem[{Elman(1990)}]{elman1990finding}
Jeffrey~L Elman. 1990.
\newblock Finding structure in time.
\newblock {\em Cognitive science\/} 14(2):179--211.

\bibitem[{Gers et~al.(2000)Gers, Schmidhuber, and Cummins}]{gers2000learning}
Felix~A Gers, J{\"u}rgen Schmidhuber, and Fred Cummins. 2000.
\newblock Learning to forget: Continual prediction with lstm.
\newblock {\em Neural computation\/} 12(10):2451--2471.

\bibitem[{Graves(2013)}]{graves2013generating}
Alex Graves. 2013.
\newblock Generating sequences with recurrent neural networks.
\newblock {\em arXiv preprint arXiv:1308.0850\/} .

\bibitem[{Hinton et~al.(2006)Hinton, Osindero, and Teh}]{hinton2006fast}
Geoffrey~E Hinton, Simon Osindero, and Yee-Whye Teh. 2006.
\newblock A fast learning algorithm for deep belief nets.
\newblock {\em Neural computation\/} 18(7):1527--1554.

\bibitem[{Hinton et~al.(2012)Hinton, Srivastava, Krizhevsky, Sutskever, and
  Salakhutdinov}]{hinton2012improving}
Geoffrey~E Hinton, Nitish Srivastava, Alex Krizhevsky, Ilya Sutskever, and
  Ruslan~R Salakhutdinov. 2012.
\newblock Improving neural networks by preventing co-adaptation of feature
  detectors.
\newblock {\em arXiv preprint arXiv:1207.0580\/} .

\bibitem[{Hochreiter and Schmidhuber(1997)}]{hochreiter1997long}
Sepp Hochreiter and J{\"u}rgen Schmidhuber. 1997.
\newblock Long short-term memory.
\newblock {\em Neural computation\/} 9(8):1735--1780.

\bibitem[{Jozefowicz et~al.(2015)Jozefowicz, Zaremba, and
  Sutskever}]{jozefowicz2015empirical}
Rafal Jozefowicz, Wojciech Zaremba, and Ilya Sutskever. 2015.
\newblock An empirical exploration of recurrent network architectures.
\newblock {\em Journal of Machine Learning Research\/} .

\bibitem[{Kiperwasser and Goldberg(2016{\natexlab{a}})}]{kiperwasser2016easy}
Eliyahu Kiperwasser and Yoav Goldberg. 2016{\natexlab{a}}.
\newblock Easy-first dependency parsing with hierarchical tree lstms.
\newblock {\em arXiv preprint arXiv:1603.00375\/} .

\bibitem[{Kiperwasser and Goldberg(2016{\natexlab{b}})}]{kiperwasser2016simple}
Eliyahu Kiperwasser and Yoav Goldberg. 2016{\natexlab{b}}.
\newblock Simple and accurate dependency parsing using bidirectional lstm
  feature representations.
\newblock {\em arXiv preprint arXiv:1603.04351\/} .

\bibitem[{Kuncoro et~al.(2016)Kuncoro, Sawai, Duh, and
  Matsumoto}]{kuncoro2016dependency}
Adhiguna Kuncoro, Yuichiro Sawai, Kevin Duh, and Yuji Matsumoto. 2016.
\newblock Dependency parsing with lstms: An empirical evaluation.
\newblock {\em arXiv preprint arXiv:1604.06529\/} .

\bibitem[{Le et~al.(2015)Le, Jaitly, and Hinton}]{le2015simple}
Quoc~V Le, Navdeep Jaitly, and Geoffrey~E Hinton. 2015.
\newblock A simple way to initialize recurrent networks of rectified linear
  units.
\newblock {\em arXiv preprint arXiv:1504.00941\/} .

\bibitem[{Marcus et~al.(1993)Marcus, Marcinkiewicz, and
  Santorini}]{marcus1993building}
Mitchell~P Marcus, Mary~Ann Marcinkiewicz, and Beatrice Santorini. 1993.
\newblock Building a large annotated corpus of english: The penn treebank.
\newblock {\em Computational linguistics\/} 19(2):313--330.

\bibitem[{Mikolov et~al.(2014)Mikolov, Joulin, Chopra, Mathieu, and
  Ranzato}]{mikolov2014learning}
Tomas Mikolov, Armand Joulin, Sumit Chopra, Michael Mathieu, and Marc'Aurelio
  Ranzato. 2014.
\newblock Learning longer memory in recurrent neural networks.
\newblock {\em arXiv preprint arXiv:1412.7753\/} .

\bibitem[{Mikolov et~al.(2010)Mikolov, Karafi{\'a}t, Burget, Cernock{\`y}, and
  Khudanpur}]{mikolov2010recurrent}
Tomas Mikolov, Martin Karafi{\'a}t, Lukas Burget, Jan Cernock{\`y}, and Sanjeev
  Khudanpur. 2010.
\newblock Recurrent neural network based language model.
\newblock In {\em Interspeech\/}. volume~2, page~3.

\bibitem[{Nair and Hinton(2010)}]{nair2010rectified}
Vinod Nair and Geoffrey~E Hinton. 2010.
\newblock Rectified linear units improve restricted boltzmann machines.
\newblock In {\em Proceedings of the 27th international conference on machine
  learning (ICML-10)\/}. pages 807--814.

\bibitem[{Nivre(2004)}]{Nivre:2004}
Joakim Nivre. 2004.
\newblock
  \href{http://dl.acm.org/citation.cfm?id=1613148.1613156}{Incrementality in
  deterministic dependency parsing}.
\newblock In {\em Proceedings of the Workshop on Incremental Parsing: Bringing
  Engineering and Cognition Together\/}. Association for Computational
  Linguistics, Stroudsburg, PA, USA, IncrementParsing '04, pages 50--57.
\newblock
  \href{http://dl.acm.org/citation.cfm?id=1613148.1613156}{http://dl.acm.org/citation.cfm?id=1613148.1613156}.

\bibitem[{Pennington et~al.(2014)Pennington, Socher, and
  Manning}]{pennington2014glove}
Jeffrey Pennington, Richard Socher, and Christopher~D. Manning. 2014.
\newblock \href{http://www.aclweb.org/anthology/D14-1162}{Glove: Global vectors
  for word representation}.
\newblock In {\em Empirical Methods in Natural Language Processing (EMNLP)\/}.
  pages 1532--1543.
\newblock
  \href{http://www.aclweb.org/anthology/D14-1162}{http://www.aclweb.org/anthology/D14-1162}.

\bibitem[{Sutskever et~al.(2013)Sutskever, Martens, Dahl, and
  Hinton}]{sutskever2013importance}
Ilya Sutskever, James Martens, George Dahl, and Geoffrey Hinton. 2013.
\newblock On the importance of initialization and momentum in deep learning.
\newblock In {\em International conference on machine learning\/}. pages
  1139--1147.

\bibitem[{Weiss et~al.(2015)Weiss, Alberti, Collins, and
  Petrov}]{weiss2015structured}
David Weiss, Chris Alberti, Michael Collins, and Slav Petrov. 2015.
\newblock Structured training for neural network transition-based parsing.
\newblock {\em arXiv preprint arXiv:1506.06158\/} .

\bibitem[{Zaremba et~al.(2014)Zaremba, Sutskever, and
  Vinyals}]{zaremba2014recurrent}
Wojciech Zaremba, Ilya Sutskever, and Oriol Vinyals. 2014.
\newblock Recurrent neural network regularization.
\newblock {\em arXiv preprint arXiv:1409.2329\/} .

\bibitem[{Zhang et~al.(2015)Zhang, Lu, and Lapata}]{zhang2015top}
Xingxing Zhang, Liang Lu, and Mirella Lapata. 2015.
\newblock Top-down tree long short-term memory networks.
\newblock {\em arXiv preprint arXiv:1511.00060\/} .

\end{thebibliography}
\bibliographystyle{acl_natbib}

\end{document}